\documentclass[sigconf]{acmart}

\usepackage{hyperref}
\usepackage{url}
\usepackage{smile}
\usepackage{dblfloatfix}
\usepackage[ruled,linesnumbered]{algorithm2e}
\usepackage{placeins}
\usepackage{hyperref}
\usepackage{graphicx}

\newcommand{\model}{CAG-ODE}

\AtBeginDocument{%
  }

\setcopyright{acmcopyright}
\copyrightyear{2024} 
\acmYear{2024} 
\setcopyright{rightsretained} 
\acmConference[WWW '24]{Proceedings of the ACM Web Conference 2024}{May 13--17, 2024}{Singapore, Singapore}
\acmBooktitle{Proceedings of the ACM Web Conference 2024 (WWW '24), May 13--17, 2024, Singapore, Singapore}\acmDOI{10.1145/3589334.3648148}
\acmISBN{979-8-4007-0171-9/24/05}

\begin{document}

\title{Causal Graph ODE: Continuous Treatment Effect Modeling in Multi-agent Dynamical Systems}

\author{Zijie Huang$\dagger$}
\authornote{Equally contributed. $\dagger$ Corresponding author.}
\affiliation{%
\department{Computer Science Department}
  \institution{University of California, Los Angeles}
  \city{Los Angeles}
  \state{California}
  \country{USA}
}
\email{zijiehuang@cs.ucla.edu}

\author{Jeehyun Hwang}
\authornotemark[1]
\affiliation{%
\department{Computer Science Department}
  \institution{University of California, Los Angeles}
  \city{Los Angeles}
  \state{California}
  \country{USA}
}
\email{hwanggh96@gmail.com}

\author{Junkai Zhang}
\authornotemark[1]
\affiliation{%
\department{Computer Science Department}
  \institution{University of California, Los Angeles}
  \city{Los Angeles}
  \state{California}
  \country{USA}
}
\email{jkzhang@g.ucla.edu}

\author{Jinwoo Baik}
\affiliation{%
\department{Computer Science Department}
  \institution{University of California, Los Angeles}
  \city{Los Angeles}
  \state{California}
  \country{USA}
}
\email{jbaik23@g.ucla.edu}

\author{Weitong Zhang}
\affiliation{%
\department{Computer Science Department}
  \institution{University of California, Los Angeles}
  \city{Los Angeles}
  \state{California}
  \country{USA}
}
\email{weightzero@g.ucla.edu}

\author{Dominik Wodarz}
\affiliation{%
\department{Department of Ecology, Behavior \& Evolution}
  \institution{University of California, San Diego}
  \city{San Diego}
  \state{California}
  \country{USA}
}
\email{dwodarz@ucsd.edu}

\author{Yizhou Sun}
\affiliation{%
\department{Computer Science Department}
  \institution{University of California, Los Angeles}
  \city{Los Angeles}
  \state{California}
  \country{USA}
}
\email{yzsun@cs.ucla.edu}

\author{Quanquan Gu}
\affiliation{%
\department{Computer Science Department}
  \institution{University of California, Los Angeles}
  \city{Los Angeles}
  \state{California}
  \country{USA}
}
\email{qgu@cs.ucla.edu}

\author{Wei Wang}
\affiliation{%
\department{Computer Science Department}
  \institution{University of California, Los Angeles}
  \city{Los Angeles}
  \state{California}
  \country{USA}
}
\email{weiwang@cs.ucla.edu}








\renewcommand{\shortauthors}{Zijie Huang et al.}

\begin{abstract}
Real-world multi-agent systems are often dynamic and continuous, where the agents co-evolve and undergo changes in their trajectories and interactions over time. For example, the COVID-19 transmission in the U.S. can be viewed as a multi-agent system, where states act as agents and daily population movements between them are interactions. Estimating the counterfactual outcomes in such systems enables accurate future predictions and effective decision-making, such as formulating COVID-19 policies.
However, existing methods fail to model the continuous dynamic effects of treatments on the outcome, especially when multiple treatments (e.g., "stay-at-home" and "get-vaccine" policies) are applied simultaneously.
To tackle this challenge,  we propose Causal Graph Ordinary Differential Equations (CAG-ODE), a novel model that captures the continuous interaction among agents using a Graph Neural Network (GNN) as the ODE function. The key innovation of our model is to learn time-dependent representations of treatments and incorporate them into the ODE function, enabling precise predictions of potential outcomes. To mitigate confounding bias, we further propose two domain adversarial learning-based objectives, which enable our model to learn balanced continuous representations that are not affected by treatments or interference. Experiments on two datasets (i.e., COVID-19 and tumor growth) demonstrate the superior performance of our proposed model. \footnote{Our code implementation can be found at \url{https://github.com/Jun-Kai-Zhang/CAG-ODE.git}.}
\end{abstract}




\maketitle

\allowdisplaybreaks

\section{Introduction}
Many real-world multi-agent systems are dynamic and continuous, where agents (nodes) interact and exhibit complex behaviors over time. This results in time-evolving node trajectories and dynamic interaction edges. An example is the spread of  COVID-19 in the U.S., where states act as agents and daily migration patterns across states form interaction edges~\cite{huang2021coupled,covidpolicy}. 
Estimating the counterfactual outcomes over time in such 
systems are crucial for various applications, such as formulating effective policies and designing medical treatment plans~\cite{seedat2022continuous,bica2020crn,bellot2021policy}. 
This can achieve more accurate predictions than non-causal methods by considering the influence of biased confounders. 
Confounders are variables that have influences on treatments and outcomes. 
For example, the health status of the residents in each state (confounders) can impact their level of adherence to the state's policies (treatments), which can influence future confirmed cases/deaths (outcomes). 
Non-causal methods only learn the statistical associations between treatments and outcomes from observational data, which can have non-uniform treatment distributions across confounder values, potentially leading to incorrect predictions such as taking vaccines can increase the number of confirmed cases for each state. Furthermore, causal inference for multi-agent dynamical systems enables effective decision-making by addressing causal questions such as "What if we remove a policy at a specific time" or "What if we change the order of different policies". Therefore, it serves as a promising tool for policymakers. 

Traditionally, the standard approach for causal inference over time is randomized controlled trials (RCTs)~\cite{chalmers1981method}, which can be very costly to obtain and can raise some ethical problems~\cite{seedat2022continuous,bica2020crn}. Thus, researchers have turned to using observational data and employed methods like 
linear regression~\cite{robins2000marginal}, recurrent neural networks (RNNs)~\cite{lim2018forecasting,BicaAJS20}, and Transformers~\cite{MelnychukFF22} to estimate counterfactual outcomes with time dependencies. 
However, causal inference for multi-agent dynamical systems presents unique challenges. 

One is that most existing methods~\cite{seedat2022continuous,bica2020crn} assume that nodes are independent, meaning their trajectories are determined solely by their own treatments. 
Some~\cite{anonymous2023cfgode} considers the influence of neighboring nodes but only assumes static interactions among them, which fails to capture situations such as daily population travel patterns between states in the context of COVID-19. 

In casual terms,  influences of neighboring nodes can be categorized into two parts: 1.) time-dependent neighborhood confounding, where a node’s treatment and outcome may be confounded by the covariates of its neighbors.  For example, if cases in neighboring states rise (covariate), a state may implement a vaccine policy (treatment) that affects future confirmed cases/deaths (outcome). 2.) time-dependent interference, where the outcome of a node can be influenced by the treatments of its neighbors. For example, a state may have reduced future cases/deaths (outcome) if neighboring states have implemented a vaccine policy (covariates), as higher vaccination rates within the population flow network give stronger protection.
As the interaction edges evolve along with node trajectories, the challenges lie in predicting the neighbors of each node (edges) and then addressing the time-dependent neighborhood confounding and interference issues.

Another challenge is that current methods lack the ability to capture the continuous and dynamic effects of multiple treatments on such systems. For instance, the impact of a "stay-at-home" policy may be most significant during its initial implementation, and when a "get-vaccine" policy is subsequently introduced, the combined effect of these policies can result in a different outcome. Existing studies often focus on a single treatment~\cite{anonymous2023cfgode,seedat2022continuous} or simply append fixed multi-hot treatment representations when a node receives them. These fixed treatment representations fail to differentiate the influences of the same treatment administered at different times.

To tackle these challenges, we propose a novel causal inference framework: the \textbf{Ca}usal \textbf{G}raph \textbf{O}rdinary \textbf{D}ifferential \textbf{E}quations (CAG-ODE) to estimate the continuous counterfactual outcome of a multi-agent dynamical system in the presence of multiple treatments and time-varying confounding and interference. Building upon the recent success of graph ordinary differential equations (ODE) in capturing the continuous interaction among agents~\cite{huang2021coupled,luo2023care,LG-ODE,huang2024tango}, our key innovation is to learn time-dependent representations of simultaneous treatments and incorporate them into the ODE function to accurately account for their casual effects on the system. As nodes and edges are jointly evolving, we utilize two coupled treatment-induced ODE functions to account for their respective dynamics. To mitigate confounding bias, we further design two adversarial learning losses, which enable our model to learn balanced continuous trajectory representations unaffected by treatments or interference. Experiments on both real and simulated datasets demonstrate the effectiveness of our proposed model. The primary contributions of this paper can be summarized as follows:
\begin{itemize}
    \item We propose CAG-ODE to estimate continuous counterfactual outcomes in multi-agent systems with evolving interaction edges and multiple treatments.
    \item CAG-ODE features a novel treatment fusing module that can capture the dynamic effects of treatment over time and the combined effect of multiple treatments.
    \item Our method achieves the state-of-art results in counterfactual estimation across varying systems, and can serve as a promising tool for policymakers.
\end{itemize}

\section{Preliminaries and Related Work}

\subsection{Graph Neural Networks (GNNs)}
Graph Neural Networks (GNNs) are a class of neural networks that operate on graph-structured data by passing local messages~\cite{GCN,GAT,GIN}.
They have been extensively employed in various applications such as node classification, link prediction, and recommendation systems~\cite{huang2023concept2box,SSAGA}.
GNNs have shown to be efficient for approximating pair-wise node interactions and achieved accurate predictions for multi-agent dynamical systems \cite{NRI,jure}. 
The majority of existing studies propose discrete GNN-based simulators where they take the node features at time $t$ as input to predict the node features at time $t$+1. To further capture the long-term temporal dependency for predicting future trajectories, some work utilizes recurrent neural networks such as RNN, LSTM, or self-attention mechanism to make predictions at time $t$ +1 based on the historical trajectory sequence~\cite{VGRNN,Dygnn,hgt}. However, they restrict themselves to learning a one-step state transition function. Therefore, when we successively apply these one-step simulators to previous predictions in order to generate the rollout trajectories, error accumulates and impairs the prediction accuracy, especially for long-range prediction.

\subsection{Graph Ordinary Differential Equations for
Continuous Multi-agent Dynamical Systems}~\label{sec:graphode}
The dynamics of a multi-agent system can be captured by a series of nonlinear first-order ordinary differential equations (ODEs)~\cite{latentODE,LG-ODE,huang2023generalizing, hope}, which describe how the states of $N$ dependent variables co-evolve over continuous time: $\dot{\bm{z}}_{i}^{t}:=\frac{d \bm{z}_{i}^{t}}{d t}=g\left(\bm{z}_{1}^{t}, \bm{z}_{2}^{t} \cdots \bm{z}_{N}^{t}\right)$. Here $\bm{z}_i^t\in\mathbb{R}^d$ denotes the state variable for agent $i$ at timestamp $t$ and $g$ denotes the ODE function that drives the system to move forward. Given the initial states $\bm{z}_{1}^{0}, 
\cdots \bm{z}_{N}^{0}$ for all agents and the ODE function $g$, a numerical ODE solver such as Runge-Kutta~\cite{solver} can be used to 
evaluate $\bm{z}_i^T$ at any desired time $T$ using Eqn~\eqref{eq:ode}: 
\begin{equation}
    \bm{z}_{i}^{T}=\bm{z}_{i}^{0}+\int_{t=0}^{T} g\left(\bm{z}_{1}^{t}, \bm{z}_{2}^{t} \cdots \bm{z}_{N}^{t}\right) \mathrm d t.
    \label{eq:ode}
\end{equation}
To model the interactions among agents, recent studies~\cite{LG-ODE,huang2021coupled,ndcn,poli2019graph} propose using a GNN as the ODE function $g$ which is learned from observational data. Such GraphODE framework follows an encoder-processor-decoder architecture. The encoder computes latent initial states for all agents based on historical observations. The GNN-based ODE function then predicts the latent trajectories starting from the learned initial states. Finally, a decoder extracts the predicted dynamic features. To regularize the generated trajectories, GraphODE frameworks often adopt a variational autoencoder (VAE) structure~\cite{VAE}, where the encoder samples initial states from approximated posterior distributions. GraphODEs are promising in making long-range predictions and can handle irregularly-sampled observations effectively~\cite{LG-ODE,ndcn}.

\subsection{Causal Inference Over Time}
Time-dependent causal inference methods mainly differ in how they deal with confounding. They differ from traditional statistical time series analysis~\cite{kipf2018neural,zhang2023crossformer,bai2022temporal} which we do not discuss in this paper.  Traditionally, many statistical tools that are applied, such as marginal structural models (MSMs)~\cite{robins2000marginal} utilize the inverse probability of treatment weighting (IPTW). Recently,  representation learning-based balancing approaches are proposed, which learn representations that are not predictable of the treatments to ensure unbiased outcome prediction~\cite{BicaAJS20,MelnychukFF22}. 
However, one major limitation is that they are discrete methods, which can offer poor performance on continuous systems such as the spread of COVID-19. 
There are a series of works~\cite{seedat2022continuous,bica2020crn,gwak2020neural,de2022predicting} that estimate the continuous counterfactual outcomes through neural ODEs or neural controlled differential equations (CDEs). Despite their success, they assume that nodes are independent of each other, regardless of their interactions. One recent work~\cite{anonymous2023cfgode} proposed to parameterize the ODE function with a GNN for multi-agent settings. However, this model cannot handle evolving graph structures and the effect of multiple treatments.

\section{Problem Definition}\label{sec:problem}
We consider a dynamical system of $N$ agents as an evolving interaction graph $\mathcal{G}^t = \{\mathcal{V},\mathcal{E}^t\}$, where nodes $\mathcal{V} = \{v_1, v_2, \cdots, v_N\}$ are agents and $\mathcal{E}^t$ are the weighted edges among them, denoting agents' dynamic interaction that changes over time. Each node is associated with time-varying causal characteristics, which we introduce in the following along with the casual inference framework.

We follow the longitudinal causal inference setting for predicting future potential outcomes as in~\cite{rubin1978bayesian}.  We denote the observational data at timestamp $t$ as $(\mathbf{X}^t, \mathbf{W}^t, \mathbf{A}^t, \mathbf{Y}^t)$, where $\mathbf{X}^t\in\mathbb{R}^{N\times d_1}$ represents the time-varying covariates (e.g., the health status of residents) of $N$ agents. $\mathbf{W}^t \in \mathbb{R}^{N\times N}$ represents the weighted adjacency matrix, whose element $w_{i\rightarrow j}\in \mathbb{R}$ is the weight of the directed edge that points from node $i$ to node $j$ and may be asymmetric. $\mathbf{A}^t\in \{0,1\}^{N\times K}$ are time-dependent treatments, where $\mathbf{A}_{kj}^t = 1$ denotes the $k^{th}$ treatment assigned to node $i$ at timestamp $t$, and $K$ is the number of heterogeneous treatments.  $\mathbf{Y}^t\in\mathbb{R}^{N\times d_2}$ is the time-dependent outcome, such as the number of confirmed cases in each state, which can be part of $\mathbf{X}^t$. The historical observations up to time $t$ is represented as $\mathcal{H}^t=\left\{\overline{\mathbf{X}}^t, \overline{\mathbf{W}}^t, \overline{\mathbf{A}}^t, \overline{\mathbf{Y}}^t\right\}$, where $\overline{\mathbf{X}}^t, \overline{\mathbf{W}}^t, \overline{\mathbf{A}}^t, \overline{\mathbf{Y}}^t$ contain all $\mathbf{X}^{t^-},\mathbf{W}^{t^-},\mathbf{A}^{t^-},\mathbf{Y}^{t^-}\left(t^{-} \leq t\right)$. 
We aim to predict the unbiased potential outcomes $\mathbb{E}\big(\mathbf{Y}^{t^{+}}\big(\mathbf{A}^{t^{+}}=a\big)|\mathcal{H}^t\big)$ under any treatment assignment $a$\footnote{The potential outcome can also be formalized using \textit{do} operation~\cite{pearl2009causality}}. Here, $a$ is the dynamic treatment trajectory (e.g. sequences of state policies). As only one of the potential outcome trajectories is observed for each treatment assignment, we refer to the unobserved potential outcomes as counterfactuals~\cite{BicaAJS20,seedat2022continuous}.

To make potential outcomes identifiable from observational data, we follow three standard assumptions~\cite{BicaAJS20,seedat2022continuous,anonymous2023cfgode} below:

\textbf{Assumption 1: Consistency}. The potential outcome is equal to the observed factual outcome if $\mathbf{A}^{t} = a^{t} $: $\mathbf{Y}^{t^+}(\mathbf{A}^{t}=a^{t}) = \mathbf{Y}^{t^+}$.

\textbf{Assumption 2: Overlap}. At any time point $t^+$, there is some positive probability of treatment assignment regardless of the historical observation: $0 < P(\mathbf{A}^{t^+}=a\mid \mathcal{H}^t) < 1,~\forall \mathcal{H}^t,~t<t^+$.

The last assumption defines unconfoundedness (strong ignorability) in  dynamical systems. 
We first define the interference effects caused by neighbors' treatments of node $i$ as $\mathbf{G}_{i}^t =\sum_{j\in \mathcal{N}_i}\frac{1}{|N_i|} \mathbf{A}_j^t\in \mathbb{R}^{K}$, which is the proportion of treated nodes in node $i$'s neighbors for each treatment type. 
We refer to $\mathbf{G}_i^t$ as interference summary, which assumes that a node is only influenced by treatments of its immediate neighbors as in previous studies~\cite{anonymous2023cfgode,ma2022learning, jiang2022estimating}.

\textbf{Assumption 3: Strong Ignorability for Multi-Agent Dynamical Systems}. Given the historical observations, the potential outcome trajectory is independent of the treatments and interference summary: $\mathbf{Y}^{t^+}(\mathbf{A}^{t}=a) \perp \mathbf{A}^{t^+}, \mathbf{G}^{t^+} \mid \mathcal{H}^t,~\forall a, t$.

It ensures that it is sufficient to only condition on the historical observations and graph sequences up to $t$ to block all backdoor paths so as to estimate the potential outcome in the future. With these three assumptions, the potential outcome trajectory can be identified as: 
\begin{align*}
    \mathbb{E}\left(\mathbf{Y}^{t^+}(\mathbf{A}^{t}=a)\mid \mathcal{H}^t\right) = \mathbb{E}\left(\mathbf{Y}^{t^+}\mid \mathbf{A}^{t^+}, \mathbf{G}^{t^+}, \mathcal{H}^t\right).
\end{align*}
This enables us to estimate the potential outcomes by training a machine learning model using observational data, and to use the same model to predict counterfactual outcomes given new treatment trajectories.

\begin{figure*}[htbp]
 \centering
 \includegraphics[width=0.9\textwidth]{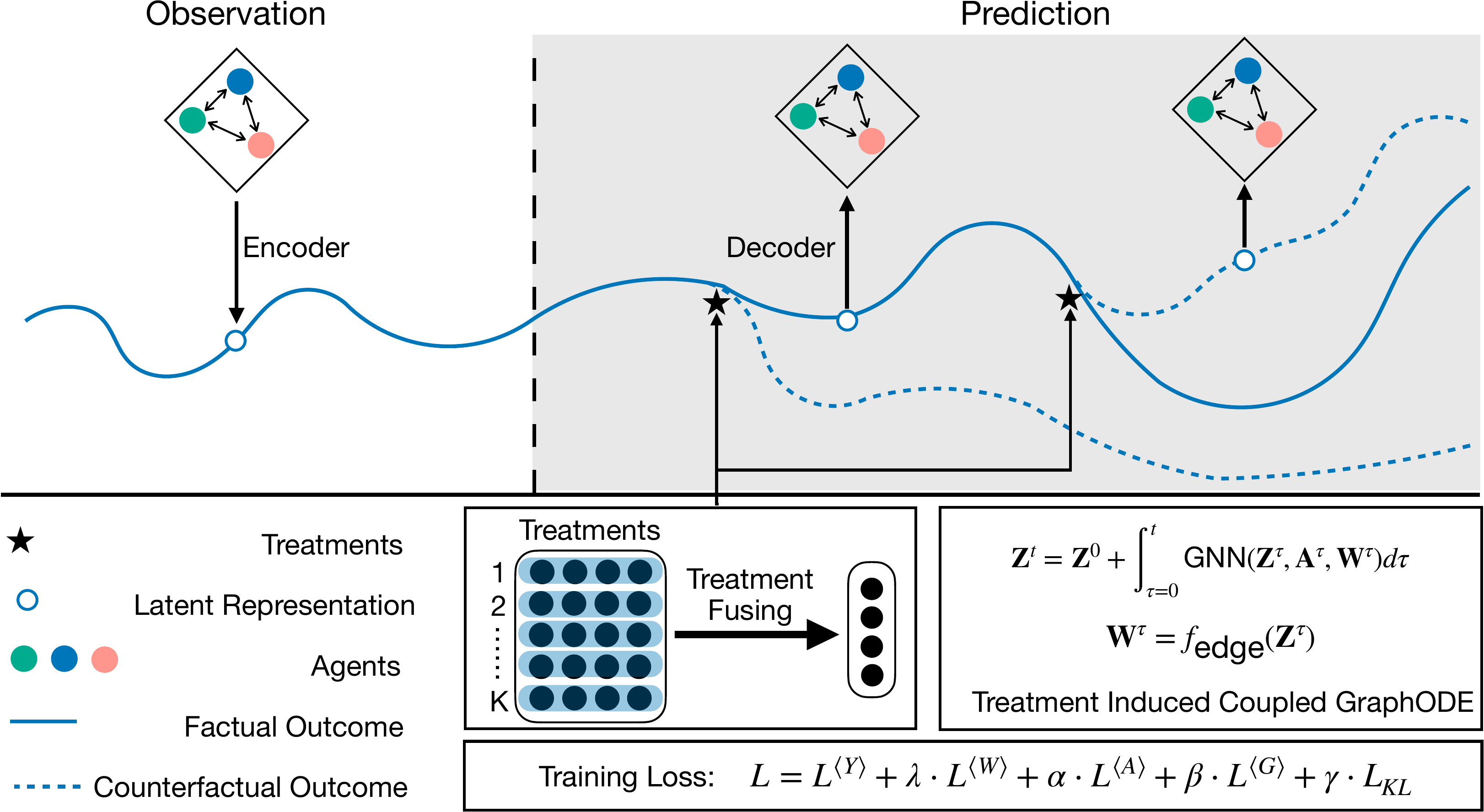}
 \vspace{-1em}
  \caption{Overall Framework of \model. The encoder first computes the latent initial states. Then the treatment-induced coupled ODE functions predict the continuous trajectories over time. Treatment representations learned through the fusing module are incorporated into the ODE functions to enable counterfactual prediction. Finally, the decoder outputs the predicted dynamics. Treatment and interference balancing losses are designed to ensure unbiased counterfactual predictions.}\label{fig:framework}
   \vspace{-1em}
\end{figure*}

\section{The Proposed Model: \model}

In this section, we present Causal Graph ODE (\model) to predict continuous counterfactual outcomes for multi-agent dynamical systems with evolving interaction edges and dynamic multi-treatment effects. 
Following the framework of GraphODEs~\cite{LG-ODE,huang2021coupled,anonymous2023cfgode,ndcn,poli2019graph}, \model~adopts the encoder-ODE generative model-decoder architecture described in Sec.~\ref{sec:graphode} to capture the continuous interaction among agents. As nodes and edges are jointly evolving, we utilize two coupled ODE functions~\cite{huang2021coupled} for the evolution of nodes and edges respectively. Contrary to GraphODEs, \model~ can perform causal reasoning by injecting treatment effects into the ODE functions, which we call {\it treatment-induced coupled graph ODE}. The multi-treatment effects are captured by a novel treatment fusing module that assigns temporal weights to the treatments using an attention mechanism.
As time-dependent confounders can result in a biased distribution of treatment assignments and imbalanced interferences due to the evolving graph structure, \model~utilizes two adversarial learning losses to ensure unbiased estimations of counterfactual outcomes. The overall framework is depicted in Figure~\ref{fig:framework}. We now discuss each module in detail.

\subsection{Spatial-Temporal Initial State Encoder}

The encoder of~\model~infers the posterior distributions  from the historical observations and then samples the latent initial states from them. It follows the architecture described in \cite{huang2021coupled}. As the evolution of different nodes is mutually influenced, we calculate the initial states for all nodes simultaneously considering their interactions over time. The initial states of edges are derived from the initial states of nodes.

\textbf{Dynamic Node Representation Learning.} We construct a graph to represent the spatial-temporal structure of multi-agent dynamical systems, with each node corresponding to an agent's observation at a particular timestamp.
There are two types of edges: spatial edges at the same timestamp and temporal edges across different timestamps. The spatial edges are formed according to the adjacency matrices, denoted as $w_{i(t)\rightarrow j(t)}$. For the temporal edges, we only consider edges from an agent's own previous observations to later observations, denoted as $w_{i(t)\rightarrow i(t')}$, where~$t' = t+1$.

The latent representations of observations are learned from this spatial-temporal graph through an attention mechanism approach. The propagation among $L$ GNN layers is depicted in Equation\eqref{eq:encode_attention}.
\begin{align}\label{eq:encode_attention}
\bm{h}_{i\left(t^{\prime}\right)}^l&=\bm{h}_{i\left(t^{\prime}\right)}^l+\sigma\left(\sum_{j(t) \in \mathcal{N}_{i\left(t^{\prime}\right)}} e_{j(t) \rightarrow i\left(t^{\prime}\right)}^l \times \bm{W}_v \hat{\bm{h}}_{j(t)}^{l-1}\right),\nonumber \\
e_{j(t) \rightarrow i\left(t^{\prime}\right)}^l&=w_{j(t) \rightarrow i\left(t^{\prime}\right)} \times \alpha_{j(t) \rightarrow i\left(t^{\prime}\right)}^l, \nonumber \\ 
\alpha_{j(t) \rightarrow i\left(t^{\prime}\right)}^l&=\left(\bm{W}_k \hat{\bm{h}}_{j(t)}^{l-1}\right)^T\left(\bm{W}_q \bm{h}_{i\left(t^{\prime}\right)}^{l-1}\right) \cdot \frac{1}{\sqrt{d}}, \\ 
\hat{\bm{h}}_{j(t)}^{l-1}&=\bm{h}_{j(t)}^{l-1}+\mathrm{TE}\left(t-t^{\prime}\right), \nonumber \\ 
\operatorname{TE}(\Delta t)_{2 i}&=\sin \left(\frac{\Delta t}{10000^{2 i / d}}\right), \operatorname{TE}(\Delta t)_{2 i+1}=\cos \left(\frac{\Delta t}{10000^{2 i / d}}\right).  \nonumber   
\end{align}
Here, $\bm{h}_{i(t)}^{l}$ represents the agent $i$ at time $t$ from layer $l$. The attention score $e_{j(t) \rightarrow i\left(t^{\prime}\right)}^l$ is defined as the product of edge weights $w_{j(t) \rightarrow i\left(t^{\prime}\right)}$ and affinity score $\alpha_{j(t) \rightarrow i\left(t^{\prime}\right)}^l$, which is computed using the representations of the sender and receiver nodes. Additionally, we incorporate temporal embedding, denoted as $\mathrm{TE}$, into the sender node's representation to establish temporal distinction. Then, the final representation is obtained from the $L$ layer as $\bm{h}_{i(t)} = \bm{h}^L_{i(t)}$.

\textbf{Sequence Representation Learning.} Then, we employ self-attention to compute the sequence representation of observed temporal information for each node, where $\hat{\bm{h}}_{i(t)} = \bm{h}_{i(t)} + \text{TE}(t)$.
\begin{equation}
    \bm{u}_i = \frac{1}{N} \sum_{t=1}^T (\bm{a}_i^T \hat{\bm{h}}_{i(t)} \hat{\bm{h}}_{i(t)}), \bm{a}_i = \text{tanh}\left(\left(\frac{1}{N} \sum_{t=1}^T \hat{\bm{h}}_{i(t)}  \right)\bm{W}_a\right).
\end{equation}

Finally, the mean and variance of the posterior distribution is obtained through a neural network $f_{\text{ddist}}$ from the sequence representation $\bm{u}_i$.
\begin{align*}
    \bz_i^0\sim q_\phi(\bz_i^0 | \cH^0)=\cN(\bmu_{z_i^0}, \sigma_{\bz_i^0}^2),~ \bmu_{\bz_i^0},\sigma_{\bz_i^0} = f_{\text{dist}}(\bu_i).
\end{align*}

Next, the latent initial state for an edge is given by $\bz_{i\rightarrow j}^0 = f_{\text{edge}}([\bz_i^0, \bz_j^0])$,
where $f_{\text{edge}}$ is parameterized by a neural network and $[,]$ is concatenation operation.

\subsection{Treatment Fusing}
To conduct causal inference with~\model, we propose to inject the dynamic effects of multiple treatments into the ODE function. Treatments can have time-varying effects in multi-agent dynamical systems and they can occur simultaneously, resulting in a combined effect.  To model such complex behaviors, we propose a novel treatment fusing module that assigns temporal weights to multiple treatments through an attention mechanism. The temporal weight of treatment at timestamp $t$ is dependent on both the start time of each treatment and the occurrence of other treatments as shown in Eqn~\eqref{eq:treatment_representation}. Let $\bm{e}_k\in\mathbb{R}^K$ be the one-hot representation of treatment $k$. We first add it with the temporal encoding TE\cite{huang2021coupled, attention} to account for the time elapsed since the start of the treatment $t^\prime$. Here $\mathbf{A}_{ik}^{t}\in \{0,1\}$ is an indicator showing whether treatment $k$ would be applied to agent $i$ at timestamp $t$. Therefore the computed treatment representation $\hat{\bo}_{ik}^{t}$ becomes zero when $\mathbf{A}_{ik}^{t} = 0$, to ensure computational efficiency. A contraction matrix $\mathbf{W}_q$ is then used to transform this sparse representation into a more compact form.

\begin{equation}\label{eq:treatment_representation}
\begin{aligned}
& \hat{\bo}_{ik}^{t}=\mathbf{A}_{ik}^t \be_k+\mathrm{TE}\left(t-t^{\prime}\right)\ind[\mathbf{A}_{ik}^t = 1],\quad \bo_{ik}^{t}  = \mathbf{W}_q \hat{\bo}_{ik}^{t}, \\
& \operatorname{TE}(\Delta t)_{2 i}=\sin \left(\Delta t / M^{2 i / d}\right), \\
&\operatorname{TE}(\Delta t)_{2 i+1}=\cos \left(\Delta t / M^{2 i / d}\right),M=10000.
\end{aligned}
\end{equation}

To account for the combined effect of simultaneous treatments, we compute the combined treatment representation as a weighted sum of all in-effect treatments at timestamp $t$ (Eqn~\ref{eq:weighted_sum}). We first compute an attention vector $m_i^t$ as the tanh-transformed average of all the treatment representations, $\hat{\bo}_{ij}^{t}$. Each treatment's weight is derived from the dot product of its representation and $m_i^t$, thereby integrating each treatment's influence into $\bo_i^t$.
\begin{equation}
\label{eq:weighted_sum}
\begin{aligned}
\bo_i^t &= \frac{1}{K} \sum_{k} \left({\bm{m}_i^t}^\top \hat{\bo}_{ik}^{t} \hat{\bo}_{ik}^{t}\right), \bm{m}_i^t = \text{tanh}\left(\left(\frac{1}{K} \sum_{k} \bo_{ik}^{t}  \right)\mathbf{W}_m\right)  .  
\end{aligned}
\end{equation}
{The fusing operation has a time complexity of $O(K)$ if having K treatments and therefore is able to scale up to larger systems.}

\subsection{Treatment-Induced GraphODE}
We use two coupled ODEs to predict the latent trajectories for nodes and edges respectively,  accounting for their co-evolution~\cite{huang2021coupled}. We incorporate the learned treatment representations into the ODEs to enable counterfactual predictions in the future.
Specifically, the co-evolution of nodes and edges is depicted in Eqn~\ref{eq:ODEs}. 
{The co-evolution depends on all historical information implicitly as $\Zb_t$ embeds the trajectories up to time $t$.} $\tilde{\mathbf{W}}_A^t=\mathbf{D}^{-1} \mathbf{W}_A^t$ is the normalized adjacency matrix and $\mathbf{D}$ is the diagonal degree matrix defined as $\mathbf{D}_{ii} = \sum_j {\mathbf{W}_A^t}_{ij}$. $f_e,f_{\text{self}},f_{\text {edge2value }}$ are all implemented as Multi-Layer Perceptrons (MLPs). To incorporate the treatment effect into the function, we use a linear transformation $\mathbf{W} $ to merge the latent states of nodes $\mathbf{Z}^t$ and the treatment representation $\bm{O}^t$. In this way, the latent trajectories of agents are affected not only by their own past trajectories and treatments but also by the trajectories and treatments of their interacting agents.
\begin{equation}\label{eq:ODEs}
\begin{aligned}
&\frac{\mathrm d \mathbf{Z}^t}{\mathrm d t}=\sigma\left(\tilde{\mathbf{W}}_A^t \mathbf{W}[\mathbf{Z}^t,\mathbf{O}^t]\right)-\mathbf{Z}^t+\mathbf{Z}^0,\\
& \frac{\mathrm d \bz_{i \rightarrow j}^t}{\mathrm d t}=f_e\left(\left[\bz_i^t, \bz_j^t\right]\right)+f_{\text {self }}\left(\bz_{i \rightarrow j}^t\right), \\ 
&{\mathbf{W}_A^t}_{ij}=f_{\text {edge2value }}\left(\bz_{i \rightarrow j}^t\right), \quad \tilde{\mathbf{W}}_A^t=\mathbf{D}^{-1} \mathbf{W}_A^t.\end{aligned}
\end{equation}

\subsection{Outcome Prediction}
Given the treatment representations, the ODE functions, the latent initial states for nodes and edges,  and the latent trajectories for all agents can be determined using any black-box ODE solver.
 Finally, we compute the predicted trajectories for each agent and their interactions based on the decoding likelihoods in Eqn~\eqref{eq:generative}, where $f_{\text{decN}}$ and $f_{\text{decE}}$ are node and edge decoding functions respectively. They output the means of the normal distributions $ p(\bm{y}_{i}^{t} | \bm{z}_{i}^{t})$ and $p(\bm{w}_{i\rightarrow j}^{t} | \bm{z}_{i}^{t})$, which we treat as the predicted values from our model.
\begin{equation}
\begin{aligned}
     \bm{y}_i^{t} \sim p(\bm{y}_{i}^{t} | \bm{z}_{i}^{t}) = f_{\text{decN}}(\bm{z}_{i}^{t}),~\bm{w}_{i\rightarrow j}^{t} \sim p(\bm{w}_{i\rightarrow j}^{t} | \bm{z}_{i}^{t}) = f_{\text{decE}}(\bm{z}_{i}^{t}).
\end{aligned}
    \label{eq:generative}
\end{equation}
We implemented all of our decoders using two-layer fully connected neural networks. The node feature decoder's input dimension matches the latent state dimension $d$, while the output dimension is one, reflecting our outcome of interest. The edge decoder's input dimension is $2d$ and the output dimension is 1. The treatment decoder also has an input dimension equal to the latent state's dimension $d$. However, its output dimension matches the number of distinct treatments, predicting the probability of each treatment being chosen. Lastly, the interference decoder's input dimension is the sum of the latent state dimension and the treatment embedding dimension, i.e. $2d$. Its output dimension mirrors the number of treatment options. For all decoders, the latent hidden dimension is half of their respective input dimensions.

We calculate the reconstruction loss of model predictions for nodes $\hat Y_i^t$ and edges $ \hat w_{i\rightarrow j}^t $ as: 
\begin{align*}
    L^{\left< Y \right>} = \frac{1}{N}\frac{1}{T} \sum_t \|\mathbf{Y}^t - \hat{\mathbf{Y}}^t\|^2_2,~
    L^{\left< W \right>} = \frac{1}{N^2}\frac{1}{T}  \sum_t \|\mathbf{W}_{A}^t - \hat{\mathbf{W}}_{A}^t\|^2_F.
\end{align*}

\subsection{Domain Adversarial Learning}
In observational data, treatment assignments are not randomized but are biased based on time-varying confounder values. This can lead to increased variance and bias in counterfactual estimation~\cite{seedat2022continuous}. In multi-agent dynamical systems, unbalanced interference from neighboring agents further exacerbates this effect and alters the state of each agent. To obtain an unbiased counterfactual prediction, we need to ensure that the distribution of latent representation trajectories is invariant to treatments and interference~\cite{anonymous2023cfgode}. This guarantees that the treatments cannot be inferred from the latent trajectory representations and that the interference is not predictable when the treatment is combined with the latent representation.

To achieve this, we incorporate two adversarial learning losses into the optimization objective function and use gradient reversal layers for the implementation.

\textbf{Treatment Balancing} The treatment combinations $\hat{\mathbf{A}}^t$ can be predicted using  a decoder from the latent state $\bz_i^t$. Formally, $\hat{\mathbf{A}}_{i\cdot}^t = \Phi_{A}(r(\bz_i^t))$, where $\Phi_A$ is a neural network attempting to recover treatments from the latent state $\bz_i^t$, and the gradient reversal layer, denoted by $r$, reverses the sign of gradient during back-propagation. 
The treatment balancing can be expressed as the maximization of the following loss term through the construction of min-max games:
\begin{align*}
    L^{\left< A \right> } = - \frac{1}{N}\frac{1}{T}\frac{1}{K}\sum_{i=1}^N\sum_{t=1}^T\sum_{k=1}^{K} \sum_{j\in\{0,1\}} \ind[(\mathbf{A}^t_{ik} =j)]  \log ( \Phi^{j,k}_A(r(\bz_i^t))),
\end{align*}
where $\Phi^{j,k}_A$ represents the logits of $d_A(\cdot)$ for predicting $j$ on $k$-th treatment. 
Note that we achieve treatment balancing by letting the latent representations $\bm{z}_i^t$ not be predictable for each individual treatment. This is because the representation of multiple treatments is essentially a linear combination of individual treatments. If each individual treatment is not predictable based on $\bm{z}_i^t$, then it is also impossible to use such representation to predict when multiple treatments occur together.

\textbf{Interference Balancing} Similar to treatment balancing, the interference prediction can be represented as $\hat{\mathbf{G}}_i^t = \Phi_G(r([Z_i^t, A_i^t]))$, where $d_G$ denotes a neural network designed to estimate interference. As interference is a continuous variable, we employ continuous domain adversarial training to accomplish interference balancing. By incorporating a gradient reversal layer, interference balancing can be achieved by minimizing the following loss term:
\begin{align*}
    L^{\left< G\right>} = \frac{1}{N}\frac{1}{T}\frac{1}{K}\sum_{i=1}^N\sum_{t=1}^T \|\Phi_{G}(r([\bz_i^t, \bo_i^t])) - \mathbf{G}_i^t\|_2^2.
\end{align*}

\textbf{Overall Loss} The overall training objective is defined as the weighted summation of node reconstruction loss, edge reconstruction loss, treatment balancing loss, and interference balancing loss. Since we follow the VAE framework, we also incorporate a KL divergence loss to add regularization towards the sampled initial states, which is defined as: $L_{KL} = \mathrm{KL}\left[\prod_{i=1}^N q_\phi\left(\bz_i^0 \mid \cH^0\right) \| p\left(\mathbf{Z}^0\right)\right]$.
Therefore, the overall training loss is formalized as:
\begin{align*}
    L = L^{\left< Y \right>} + \lambda L^{\left< W \right>} + \alpha L^{\left< A \right> } + \beta L^{\left< G\right>} + \gamma L_{KL}.
\end{align*}

\section{Experiments}
\begin{table*}[]
\centering
\caption{Root Mean Square Error (RMSE) for factual outcome evaluation across prediction lengths (the duration for which predictions are made). For the COVID-19 dataset, we report the mean and standard deviation accuracy with multiple runs.}
 \vspace{-1em}
\begin{tabular}{l|ccc|cccc}
\toprule
\multicolumn{1}{c|}{Dataset}           & \multicolumn{3}{c|}{Covid-19}                   & \multicolumn{3}{c}{Tumor Growth}                   \\
\multicolumn{1}{c|}{Prediction Length} & 7-days & 14-days & 21-days & 14-days & 21-days & 28-days \\ \hline
CG-ODE                                 & 4063 $\pm$ 68     & 4454 $\pm$ 100    & 4659 $\pm$ 63           & 18.37          & 21.00          & 24.58          \\
TE-CDE                                 & 7999 $\pm$ 212    & 7470 $\pm$ 289    & 6832 $\pm$ 243           & 55.45          & 55.38          & 71.23          \\
COVID-POLICY                           & 4008 $\pm$ 44      & 4128 $\pm$ 60     & 3963 $\pm$ 59  & 20.07          & 25.93         &  29.29         \\ \hline
CAG-ODE                                & \textbf{3710 $\pm$ 29}     & \textbf{3925 $\pm$ 44}     & \textbf{3933 $\pm$ 40}           & \textbf{10.91} & \textbf{10.82} & \textbf{14.84}          \\
w/o $L^{\left< G \right>}$             & 3800 $\pm$ 60     & 3987 $\pm$ 40     & 3990 $\pm$ 49           & 15.57          & 16.28          & 16.62              \\
w/o $L^{\left< A \right>}$             & 3840 $\pm$ 35     & 4100 $\pm$ 53     & 4069 $\pm$ 49           & 17.90          & 14.69          & 20.19              \\
w/o $L^{\left< G \right>}$ ,$L^{\left< A \right>}$                       & 3793 $\pm$ 23     & 4089 $\pm$ 79     & 3953 $\pm$ 38           & 17.28          & 16.72          & 24.36          \\
w/o attention                          & 3867 $\pm$ 61     & 3958 $\pm$ 31      & 4256 $\pm$ 55           & 18.91          & 17.55          & 34.45          \\ 
\bottomrule
\end{tabular}%
\medskip
\label{exp:main_table}
\vspace{-2em}
\end{table*}
\subsection{Experiment Setup}
\subsubsection{Datasets and Experiment Configuration} 

We evaluate the performance of our model using two datasets: 1.) The {\bf COVID-19 dataset}, which captures the daily COVID-19 trends of  U.S. states from April.12.2020 to Dec.31.2020. The daily population flows among states are represented as dynamic edges. Treatments are state-level COVID-19 policies. We ask the model to predict the daily confirmed cases in each state.  2.) The {\bf Tumor Growth simulation dataset} \cite{geng2017prediction}, which describes the tumor growth dynamics in different regions of patients, where they may receive differing treatments. We aim to predict the tumor volumes in each region. Additional details about the datasets can be found in Appendix~\ref{sec:dataset_descriptions}.

We predict trajectory rollouts across varying lengths and use Root Mean Square Error (RMSE) as the evaluation metric. Specifically, we train our model in a sequence-to-sequence setting where we split the trajectory of each training sample into two parts $[t_1,t_K]$ and $[t_{K+1},t_T]$. We condition the model on the first part of observations and predict the second part. To fully utilize the data points within each trajectory, we generate training and validation samples by splitting each trajectory into several chunks using a sliding window. Details can be found in Appendix~\ref{sec:training_details}.

\subsubsection{Baselines and Model Variants}
We conduct a comparative analysis of our model with three baseline models:  one non-causal continuous multi-agent baseline CG-ODE~\cite{huang2021coupled}, and two causal models: TE-CDE~\cite{seedat2022continuous} and COVID-POLICY~\cite{covidpolicy}. TE-CDE~\cite{seedat2022continuous} is a causal model that employs continuous-time differential equations to capture temporal event dependencies. COVID-Policy~\cite{covidpolicy} is another causal model designed specifically for assessing the impact of public health policies on COVID-19 outcomes. To further analyze the performance of our model, we also compare variants of our model. Each variant excludes a specific component to assess its individual impact on performance. The variants include models without treatment balancing, interference balancing, both components or the attention module.

\subsubsection{Training Details}
We employ the AdamW optimizer, as proposed in the study by Loshchilov et al. \cite{loshchilovdecoupled}, to train our model. The initial learning rate is set at $\eta = 0.005$, and the batch size is set as $8$ to accommodate memory constraints.

The Graph Neural Network (GNN) used for the encoder has a singular layer with a hidden dimension of $64$. Similarly, the GNN that parameterizes the ODE function is also comprised of a single layer. The dimension of the latent state is set at $20$, and the dimension for the embedded treatments is $5$. We assign a weight of $10$ for both the treatment balancing term $\alpha$ and the interference balancing term $\beta$. Additionally, the weight designated for the edge reconstruction error $\lambda$ is set at $0.5$.

\subsection{Performance Evaluation}
We evaluate the performance of our model, CAG-ODE, as well as the baselines using Root Mean Square Error (RMSE) across different prediction lengths. The results are shown in Table~\ref{exp:main_table} and Table~\ref{exp:flip_table}, reporting the factual and counterfactual outcomes respectively. As the COVID-19 is a real-world dataset that does not have counterfactual outcomes, we evaluate only the Tumor Growth dataset in Table~\ref{exp:flip_table}. To ensure consistent comparison, we align the prediction periods of all models with weekly intervals on the COVID-19 dataset, similar to the statistical baselines derived from their official weekly submissions to the CDC, as done in \cite{huang2021coupled}. To assess the accuracy of short-term and long-term predictions, the prediction lengths for the COVID-19 and Tumor Growth datasets are set to 7, 14, 21 days and 14, 21, and 28 days, respectively. We include longer-range predictions on the Tumor-Growth dataset in Appendix~\ref{sec:appendix_tumor_results}

\textbf{Factual Outcome Predictions.} 
 Table~\ref{exp:main_table} shows that our model, CAG-ODE, consistently outperforms the baseline models across all prediction lengths for both datasets. This underscores the effectiveness of our model in capturing the dynamic interactions among objects, especially over longer time periods. Comparing our model with  TE-CDE, we observe a performance gap that highlights the benefits of incorporating interference balancing and spatial correlation in the model. Additionally, our model outperforms the COVID-POLICY model, indicating its broader generalizability across different types of data due to modeling dynamic interactions. 
 Furthermore, our model exhibits proficiency in both short-term and long-term predictions. For instance, it achieves promising results for 21-day predictions on the COVID-19 dataset and 28-day predictions on the Tumor Growth simulation dataset.  The analysis of our model variants further emphasizes the importance of each component in the model. Particularly, the model variant excluding the attention module has the weakest performance, indicating the significance of our time-embedding attention module in effectively representing the treatment.

\begin{table*}[htbp]
\caption{Root Mean Square Error (RMSE) for counterfactual Outcome evaluation on the Tumor Growth dataset with treatment flipping ratio. Treatment F.R. (\textbf{Treatment} \textbf{F}lipping \textbf{R}atio) represents the ratio of treatments that are flipped.}
 \vspace{-1em}
\begin{tabular}{l|ccc|ccc|ccc}
\toprule
\multicolumn{1}{c|}{Prediction Length} & \multicolumn{3}{c|}{14-days} & \multicolumn{3}{c|}{21-days} & \multicolumn{3}{c}{28-days} \\
\multicolumn{1}{c|}{Treatment F.R.} & 0.25 & 0.5 & 0.75   & 0.25 & 0.5 & 0.75 & 0.25 & 0.5 & 0.75 \\ \hline
TE-CDE & 95.61 & 103.2 & 100.8 & 98.65 & 103.0 & 97.93 & 118.3 & 124.0 & 121.4 \\
COVID-POLICY & 21.32 & 22.37 & 23.31 & 26.63 & 26.83 & 27.00 & 32.01 & 32.16 & 32.21        \\ \hline
CAG-ODE & \textbf{17.23} & \textbf{16.98} & \textbf{16.96} & \textbf{18.64} & \textbf{18.84} & \textbf{18.85} & \textbf{19.91} & \textbf{19.88} & \textbf{19.87} \\
w/o $L^{\left< G \right>}$ & 20.62 & 20.53 & 20.51 & 19.70 & 19.60          & 19.55          & 21.10            & 21.41           & 21.38            \\
w/o $L^{\left< A \right>}$                       & 22.17          & 22.35          & 22.35          & 20.19          & 20.10          & 20.09          & 20.83            & 21.14           & 21.15            \\
w/o $L^{\left< G \right>}$, $L^{\left< A \right>}$                                 & 19.78          & 19.75          & 19.71          & 19.34          & 19.29          & 19.27          & 21.31        & 21.40       & 21.34        \\
w/o attention                                       & 19.09          & 18.37 & 18.13 & 22.16          & 21.78          & 21.65          & 27.70        & 27.44       & 27.38        \\ \bottomrule

\end{tabular}%
\medskip
\label{exp:flip_table}
\vspace{-1em}
\end{table*}

\textbf{Counterfactual Outcome Predictions.} In the context of a multi-agent dynamical system, the total number of possible treatments for all nodes is $\bO(K \times 2N)$, making it infeasible to enumerate all treatment combinations. To assess the robustness of each model to counterfactual treatment scenarios, we perform an experiment where we randomly flip a certain percentage of observed treatments. In Table~\ref{exp:flip_table}, we evaluate the performance when 25\%, 50\%, and 75\% of all observed treatments in each experiment are randomly flipped.
The purpose of this experiment is to examine the robustness of the models to counterfactual treatment scenarios, and since CG-ODE does not incorporate causal modeling, it is excluded from this experiment.
CAG-ODE outperforms others by a wide margin across all settings.
These findings collectively demonstrate the superiority of our proposed model, CAG-ODE, in capturing the dynamics of multi-agent systems and making accurate predictions across different time horizons. We additionally include the visualization of the learned balanced latent representations in Section~\ref{sec:appendix_visual}.

\subsection{Case Study about COVID-19 Policies}

\begin{figure*}[htbp]
    \centering
    \begin{minipage}[b]{0.24\textwidth}
        \centering
        \includegraphics[width=\textwidth]{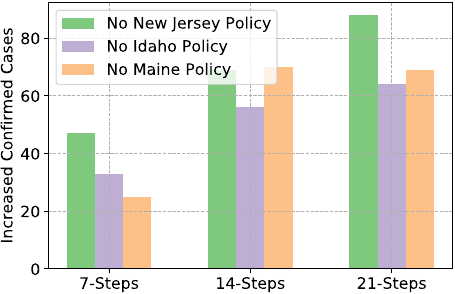}
        {(a) Remove partial states' policy.}
    \end{minipage}
    \begin{minipage}[b]{0.24\textwidth}
        \centering
        \includegraphics[width=\textwidth]{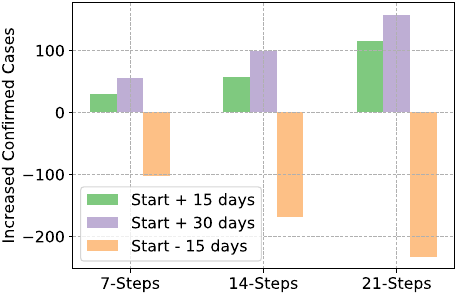}
        {(b) Change policy start date.}
    \end{minipage}
    \begin{minipage}[b]{0.24\textwidth}
        \centering
        \includegraphics[width=\textwidth]{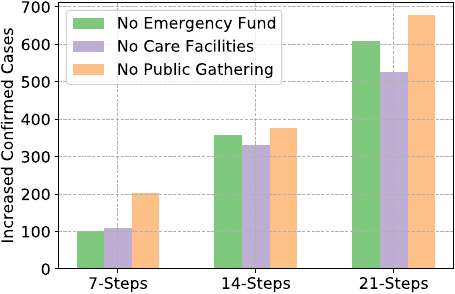}
        {(c) Remove policy across states.}
    \end{minipage}
    \begin{minipage}[b]{0.25\textwidth}
        \centering
        \includegraphics[width=\textwidth]{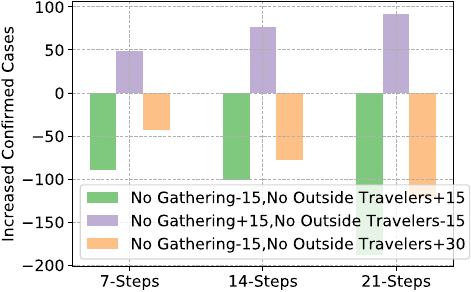}
        {(d) Change relative time of policies.}
    \end{minipage}
     \vspace{-0.5em}
    \caption{Case Study for changing different policies on the COVID-19 dataset.}
    \label{fig:case study}
    \vspace{-0.5em}
\end{figure*}

We conduct a case study to show the impact of different treatments, e.g., COVID-19 related policies, on the COVID-19 dataset as shown in Figure~\ref{fig:case study}. Specifically, we consider four different policy intervention methods and report the resulting average changes in the number of daily confirmed cases across all states in the U.S.

First, we focus on the removal of policies in three states that have the highest number of announced policies during the time frame of the COVID-19 dataset. By masking out these policies, we observe an increase in the average number of confirmed cases across states in the future. This increase is attributed to both in-state disease spread and population flow to other states. The removal of policies exacerbates the spread of COVID-19 over an extended period, as shown in Figure~\ref{fig:case study}(a), indicating that our model captures the dynamic interference resulting from agents' interactions.  

We then explore the effect of changing the starting time of a specific policy for all states. We changed the "No Public Gatherings" policy starting time for each state to be 15 days earlier, 15 and 30 days later respectively. As shown in Figure~\ref{fig:case study}(b) when announcing the policy earlier, we observe a decrease in the average number of daily confirmed cases in the future, while announcing the policy later leads to an increase. This intuitive outcome highlights the capability of our model to capture the causal relationships between policy interventions and COVID-19 spread.

Next, we analyze the impact of the top three most frequent policies across all states by removing them separately.  As shown in Figure~\ref{fig:case study}(c), the "Public Gatherings" policy has the largest effect in reducing the spread of COVID-19, even though the most frequent policy is "Emergency Funds". This demonstrates the potential of our model in assisting policymakers to identify the relative importance of each policy over time.

 \begin{figure*}[htbp]
    \centering
    \begin{minipage}[b]{0.24\textwidth}
        \centering
        \includegraphics[width=\textwidth]{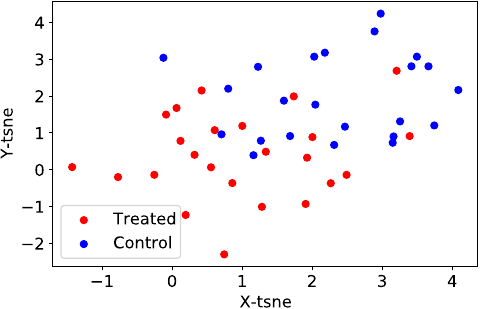}
        {(a) "State-of-Emergency" w/o. Treatment Balancing.}
    \end{minipage}
    \begin{minipage}[b]{0.24\textwidth}
        \centering
        \includegraphics[width=\textwidth]{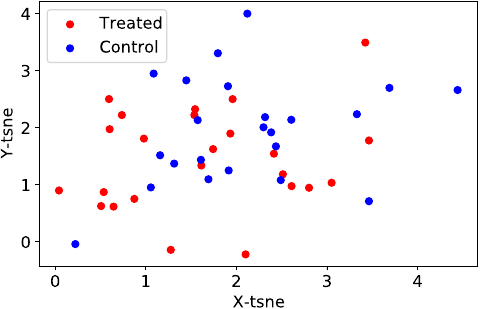}
        {(b) "State-of-Emergency" with Treatment Balancing.}
    \end{minipage}
    \begin{minipage}[b]{0.224\textwidth}
        \centering
        \includegraphics[width=\textwidth]{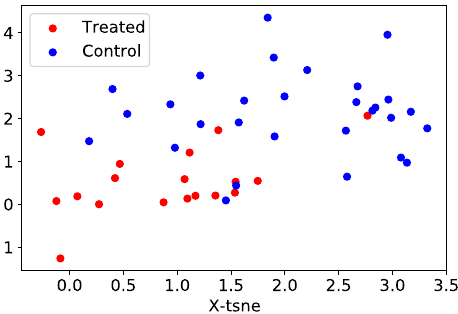}
        {(c) "No-Public-Gathering" w/o. Treatment Balancing.}
    \end{minipage}
    \begin{minipage}[b]{0.24\textwidth}
        \centering
        \includegraphics[width=\textwidth]{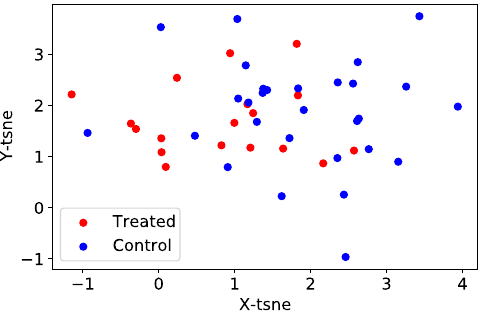}
        {(d) "No-Public-Gathering" with Treatment Balancing.}
    \end{minipage}
    \vspace{-1em}
    \caption{Treatment Balancing Visualization on the COVID-19 Dataset.}
    \label{fig:visual}
\end{figure*}

Finally, we study the effects of different orders in policy announcements, specifically focusing on the simultaneous or closely timed announcements of "No Public Gatherings" and "No Traveler from Outside States" policies. We change the announcement dates for the two policies in each state to mimic three scenarios shown in Figure~\ref{fig:case study} (d). We found that initializing the announcement of "No Public Gatherings" early generally contributes to a reduction in the spread of COVID-19 compared with "No Traveler from Outside States". We further analyzed the daily population flow during the given time frame and found that the majority of population flows are within the same states, indicating that residents of each state pose a high risk of virus transmission compared to people from other states. These insights suggest prioritizing the earlier announcement of the "No Public Gatherings" policy over the  "No Traveler from Outside States" policy can better mitigate the spread of COVID-19.

These case study results demonstrate the effectiveness of our model CAG-ODE in capturing the complex interactions between treatments, disease spread, and population flow, providing valuable insights for policymakers in making informed decisions.

\subsection{Visualization of Learned Balanced Representations}\label{sec:appendix_visual}
To further understand the effect of treatment balancing loss in~\model, we visualize the 2-D T-SNE projections of the latent representations of nodes on the COVID-19 dataset, i.e. $\bm{z}_i^t$ as shown in Figure~\ref{fig:visual}. Specifically, we visualize the latent node representations under two different treatments: "State-of-Emergency" and "No-Public-Gathering". Under each treatment (policy), we use different colors to denote whether a node receives such treatment (treated) or not (control). As shown in Figure~\ref{fig:visual}(a) and (c), the distributions of the learned representations are more distinguishable between the two groups, compared with Figure~\ref{fig:visual}(b) and (d) which have the treatment balancing loss. This indicates that~\model~ indeed learns balanced latent representations by employing the treatment balancing loss.

\section{Conclusion}
In this paper, we introduce the causal graph ODE (\model) as a model for estimating continuous counterfactual outcomes in multi-agent-dynamical systems with evolving interaction edges and dynamic multi-treatment effects. Our model builds upon existing GraphODEs and incorporates causal reasoning for multi-agent dynamical systems. We propose a novel treatment fusing module that captures the dynamic effects of multiple treatments occurring simultaneously. Through extensive experiments on both the real-world and the simulated datasets, we demonstrate the superior performance of our model across various prediction settings, validating its effectiveness. Furthermore, we leverage our model to analyze policy effects analysis on the COVID-19 dataset, providing valuable insights for policymakers.

\section{Acknowledgement}
This work was partially supported by NSF 2106859, 2200274, 2312501, NEC and NSF 2211557, NSF 1937599, NSF 2119643, NSF 2303037, NSF 20232551, NASA, SRC JUMP 2.0 Center, Cisco research grant, Picsart Gifts, and Snapchat Gifts. We would like to thank Song Jiang for his valuable discussion throughout this project.

\newpage

\bibliographystyle{ACM-Reference-Format}
\bibliography{reference}

\newpage
\appendix

\section{Appendix}
\appendix
\section{Dataset Description} \label{sec:dataset_descriptions}
\subsection{COVID-19 Dataset}\label{sec:covid19_dataset}

 Our experiments used the dataset provided by the Johns Hopkins Coronavirus Resource Center (JHU) \footnote{https://coronavirus.jhu.edu/about/how-to-use-our-data} from April 12th to December 31st, 2020. That is, we consider 264 time points, with each point representing one day. 
 The dataset contains a comprehensive range of information, but for our experiments, we focus on up to 7 specific features. These features include the daily counts of confirmed cases, deaths, recovered cases, active cases, incident rate (cases per 100,000 people), mortality rate (calculated as the number of recorded deaths multiplied by 100 divided by the number of cases), and testing rate (total test results per 100,000 people). It's worth noting that while the JHU dataset provides cumulative data for confirmed, deaths, recovered, and active cases, our experiments and models specifically consider the daily increases in these features (e.g., the number of new cases reported each day).

To capture dynamic interaction edges, we use a temporal mobility flow network among a selection of 47 states based on COVID-19 USFlows \cite{mobility}.

The treatments are represented as statewide policies that aim to combat the spread of COVID-19. We identify 58 different statewide policies enacted throughout 2020, from the data given by the Department of Health \& Human Services \footnote{https://catalog.data.gov/dataset/covid-19-state-and-county-policy-orders-9408a}. Each state enacted around 20 of these policies during the time period of April 2020 to December 2020, where the dataset provides the start and end dates of each enacted policy. In our model, treatments are encoded such that for each time point, the value is either 1 or 0 depending on whether the particular policy is enacted (for a given state) at that time or not, respectively. 

Overall, the model receives input data for a total of 264 time points, covering each of the 47 states, and includes 7 features. Alongside this, the model is also provided with treatments and a mobility graph. Prior to being used as input for our models, the data is normalized. However, when calculating the test loss for comparison with other baseline models, the output is unnormalized. Our goal is to predict either the number of confirmed cases or the number of deaths for a future period of 7, 14, or 21 days.
 
\subsection{Tumor Growth Dataset}\label{sec:tumor_growh_dataset}
We extend the state-of-the-art pharmacokinetic-pharmacodynamic (PK-PD) model of tumor growth proposed by \cite{geng2017prediction} to simulate a more complex scenario where multiple tumor regions within a single patient interact with each other. The original model characterizes patients suffering from non-small cell lung cancer and models the evolution of their tumor under the combined effects of chemotherapy and radiotherapy. For a detailed description of the original model, we refer the readers to the original paper \cite{geng2017prediction}. In our extended model, we incorporate two new terms: an interference term and a neighborhood covariate term. The volume of the tumor in region \(i\) at \(t+1\) days after diagnosis is modeled as follows:
\begin{small}
\begin{equation}
\begin{aligned}
& V_i(t + 1)  \\
= & \Bigg(1 + \underbrace{\rho \log\left(\frac{K}{V_i(t)}\right)}_{\text{Tumor Growth}} - \underbrace{\beta_c C_i(t)}_{\text{Chemotherapy}} - \underbrace{\left(\alpha_r d_i(t) + \beta_r d_i(t)^2\right)}_{\text{Radiotherapy}} + \underbrace{e_{it}}_{\text{Noise}} \Bigg) V_i(t) \nonumber \\
& + \underbrace{\left(\frac{1}{N_i} \sum_{j \in \mathcal{N}_i} \iota_c C_j(t)\right)}_{\text{Chemotherapy Interference}} + \underbrace{\left(\frac{1}{N_i} \sum_{j \in \mathcal{N}_i} \left(\iota_r d_j(t) + \iota_r d_j(t)^2\right)\right)}_{\text{Radiotherapy Interference}}  \\
&  \qquad \qquad \qquad \qquad \qquad + \underbrace{\left(\frac{1}{N_i} \sum_{j \in \mathcal{N}_i} \kappa V_j(t)\right)}_{\text{Neighborhood Covariates}}.
\end{aligned}
\label{eq:tumor_growth_eq}
\end{equation}    
\end{small}
where the parameters \(K\), \(\rho\), \(\beta_c\), \(\alpha_r\), \(\beta_r\) are sampled from the prior distributions described in \cite{geng2017prediction}, and \(e_{it} \sim N(0, 0.012)\) is a noise term that accounts for randomness in the tumor growth. The prior means for \(\beta_c\) and \(\alpha_r\) are adjusted to create three patient subgroups \(S(i) \in \{1, 2, 3\}\) as described in \cite{bica2020crn}. The chemotherapy drug concentration follows an exponential decay with a half-life of 1 day.  The time-varying confounding is introduced by modeling chemotherapy and radiotherapy assignment as Bernoulli random variables, with probabilities \(p_c\) and \(p_r\) depending on the tumor diameter. For more details, we refer the reader to the paper \cite{bica2020crn}. For our newly defined interference and neighborhood covariate terms, we set the hyperparameters \(\iota_c\) and \(\iota_r\) to 0.01, and \(\kappa\) to 0.001. These values were carefully chosen to reflect the strength of the interference and neighborhood covariates in the dataset. The number of tumors in each patient $N_i$ is fixed to 15, and for each tumor region, the number of edges connected between the tumor regions is defined randomly from the range of 22 to 45. For additional experiments shown in Appendix~\ref{sec:appendix_tumor_results}, the number of tumor regions for each patient is fixed to 5, and the number of edges ranges from 6 to 10. The dataset is input into our model similar to the COVID-19 dataset. Both chemotherapy and radiotherapy are encoded into 0 or 1 value depending on whether it was applied at a specific time point. The input data consists of 60-time points with 4 features, which include tumor volume, patient type, and the two treatments. The data is normalized for model input but unnormalized for test loss calculation. The model's objective is to predict tumor volume for future periods of 14, 21, or 28 days. We also create a longer-range dataset with 120-time points, which is described in \ref{sec:appendix_tumor_results}.

\begin{table*}[htbp]
  \centering
    \begin{tabular}{l|ccc|ccc|ccc}
    \toprule
    \multicolumn{1}{c|}{Prediction Length} & \multicolumn{3}{c|}{35-days} & \multicolumn{3}{c|}{49-days} & \multicolumn{3}{c}{63-days} \\
    \multicolumn{1}{c|}{Treatment F.R.} & 0.25 & 0.5 & 0.75   & 0.25 & 0.5 & 0.75 & 0.25 & 0.5 & 0.75 \\ \hline
    CAG-ODE & 35.00 & 34.68 & 34.50 & 34.12 & 37.92 & 37.88 & 46.71 & 47.28 & 48.12 \\
    \bottomrule
    \end{tabular}%
    \medskip
    \caption{Root Mean Square Error (RMSE) for counterfactual outcome evaluation on the longer-range Tumor Growth dataset with treatment flipping ratio: $25\%, 50\%, 75\%$.}
  \label{tab:long_range}%
\end{table*}%

\section{Data Splitting}\label{sec:training_details}
We train our model in a sequence-to-sequence setting, where we split the trajectory of each training sample into two parts $[t_1,t_K]$ and $[t_{K+1},t_T]$. We condition the model on the first part of observations and predict the second part. To fully utilize the data points within each trajectory, we generate training and validation samples by splitting each trajectory into several chunks using a sliding window with three hyperparameters: the observation length and prediction length for each sample, and the interval between two consecutive chunks (samples). We summarize the procedure in Algorithm~\ref{al:split}, where $K$ is the number of trajectories and $d$ is the input feature dimension. For both datasets, we set the observation length to be 7 and the interval to be 3. We ask the model to make predictions at varying lengths for evaluation.

\begin{algorithm}[htbp]
\caption{Data Splitting Procedure.}
\label{al:split}
\KwIn{
Original Training trajectories  {$X_{\text{input}}\in \mathbb{R}^{K\times N\times T\times d}$};\\
Observation length $O$;
Prediction length $M$;
Interval $I$;
Trajectory length $T$.\\}
\KwOut{Training samples after splitting $X_{\text{train}}$.}
sample$\_$length = $O+M$;\\
num$\_$chunk = ($T$ - sample$\_$length )//interval + 1;\\
\For{i in range (0,K)}{
    \For{j $\text{in range}$(0,num$\_$chunk,I)}
    {Generate the split training sample as $X_{\text{input}}[i,:,j:j +
\text{sample}\_\text{length},:]$\\
Add the training sample to the training set $X_{\text{train}}$.}
    }
\end{algorithm}

\section{Longer-range Prediction for the Tumor Growth Dataset}\label{sec:appendix_tumor_results}

In Table~\ref{tab:long_range}, we evaluate the performance of our model. An extended version of the simulation dataset with a range of 120 days was used in the experiment. The considered prediction lengths are 35, 49, and 63 days.

As anticipated, the prediction errors exhibit a moderate increase with longer prediction lengths, as the added duration poses a greater challenge for accurate predictions.  Note that the prediction error remains relatively stable when the treatment flipping ratio is increased from 25\% to 75\%. This observation suggests that the utilization of treatment balancing and interference balancing techniques effectively mitigates the risk of overfitting to confounding factors, ensuring CAG-ODE's robustness.

\section{Model Implementation Details}
We use the fourth-order Runge-Kutta method from the torchdiffeq python package~\cite{ode} as the ODE solver, for solving the ODE systems on a time grid that is five times denser than the observed time points. We also utilize the Adjoint method described in ~\cite{ode} to reduce memory use.

\section{Limitations}
One limitation of~\model~is that when inferring the future trajectories of nodes, we simply assume that all nodes are connected and jointly infer such edge evolution. This would bring huge computational costs when generalized to large-scale dynamical systems. In the future, we will consider more efficient sampling methods to accelerate the edge inference procedure to scale up our model. Another line of future work would be how to model more complex multiple treatment effects, including competing, hierarchical relationships.

\section{Broader Impacts}
Our work significantly enhances the performance of causal inference over multi-agent dynamical systems, which can potentially benefit a wide range of fields including public health,  biology, physics, and robotics. Our work also advances the recent study of continuous graphODE for modeling multi-agent system dynamics, providing an efficient tool for further research on AI for science.

\end{document}